\pdfoutput=1
\documentclass[11pt]{article}

\usepackage[final]{acl}

\usepackage{times}
\usepackage{latexsym}

\usepackage[T1]{fontenc}
\usepackage[utf8]{inputenc}

\usepackage{microtype}

\usepackage{inconsolata}

\usepackage{graphicx}

\usepackage{graphicx}
\usepackage{amsmath}
\usepackage{amssymb}
\usepackage{booktabs}
\usepackage{amsthm}
\usepackage{multirow}
\usepackage{makecell}
\usepackage{subfig}
\usepackage{subcaption}
\usepackage{bm}
\usepackage{array}
\usepackage{booktabs}
\usepackage{wasysym}
\usepackage{wrapfig}
\usepackage{caption}
\usepackage{algpseudocode}
\usepackage{algorithmicx,algorithm}
\usepackage{tikz}
\usepackage{tcolorbox}
\newcommand{\fakeparagraph}[1]{\noindent\textbf{#1}
}

\usepackage{cleveref}

\newtheorem{theorem}{Theorem}

\newcommand{\tool}
{\texttt{VLMInferSlow}}

\title{VLMInferSlow: Evaluating the Efficiency Robustness of Large Vision-Language Models as a Service}

\author{
	Xiasi Wang$^{1,}$\thanks{\ \ Equal contribution},  Tianliang Yao$^{2,}$\footnotemark[1], Simin Chen$^{2}$, Runqi Wang$^{3}$, Lei Ye$^{4}$ \\
    \bf Kuofeng Gao$^{5}$, \bf Yi Huang$^{6}$, \bf Yuan Yao$^{1}$ \\
$^1$The Hong Kong University of Science and Technology \quad 
$^2$Tongji University \\
$^3$Beijing Jiaotong University \quad 
$^4$Huawei \quad
$^5$Tsinghua University \\
$^6$University of Chinese Academy of Sciences \\
}

\begin{document}
\maketitle

\begin{abstract}

Vision-Language Models (VLMs) have demonstrated great potential in real-world applications. While existing research primarily focuses on improving their accuracy, the efficiency remains underexplored. Given the real-time demands of many applications and the high inference overhead of VLMs, efficiency robustness is a critical issue.  However, previous studies evaluate efficiency robustness under unrealistic assumptions, requiring access to the model architecture and parameters---an impractical scenario in ML-as-a-service settings, where VLMs are deployed via inference APIs.
To address this gap, we propose \tool, a novel approach for evaluating VLM efficiency robustness in a realistic black-box setting. \tool\ incorporates fine-grained efficiency modeling tailored to VLM inference and leverages zero-order optimization to search for adversarial examples.   
Experimental results show that \tool\ generates adversarial images with imperceptible perturbations, increasing the computational cost by up to 128.47\%. We hope this research raises the community's awareness about the efficiency robustness of VLMs.
\footnote{Code is at \href{https://github.com/wangdaha1/VLMInferSlow}{https://github.com/wangdaha1/VLMInferSlow}. Emails: Xiasi Wang: xwangfy@connect.ust.hk, Tianliang Yao: yaotianliang@tongji.edu.cn, Simin Chen: 1152705@tongji.edu.cn, Runqi Wang: rqwang@bjtu.edu.cn, Lei Ye: yeplewis@gmail.com, Kuofeng Gao: gkf21@mails.tsinghua.edu.cn, Yi Huang: yi.huang@siat.ac.cn, Yuan Yao: yuany@ust.hk (corresponding author).}

\end{abstract}

% \TODO{github link} 

\section{Introduction}

% \TODO{VLMs are important and popular}

Large vision-language models (VLMs) have recently achieved impressive performance across a wide range of multi-modal tasks, including image captioning, visual question answering, and visual reasoning~\cite{li2022blip, alayrac2022flamingo}. 
The success of these models is driven by their underlying billions of parameters, which require substantial computational resources for effective deployment~\cite{de2023growing}.

% \TODO{necessary of efficiency of VLMs}

When deploying VLMs in real-world applications, inference efficiency is a critical concern. For example, applications like Microsoft's Seeing AI~\cite{seeingai} and Be My Eyes~\cite{bemyeye} depend on VLMs to deliver real-time object descriptions for individuals with visual impairments. If these models fail to provide instant feedback, users may face safety risks in critical situations.
In addition to meeting real-time performance requirements, energy efficiency is also a critical factor. Both NVIDIA and Amazon Web Services report that the inference phase during deployment accounts for over 90\% of the total machine learning energy consumption, highlighting the significance of inference efficiency of these VLM applications~\cite{patterson2021carbon}.

\begin{figure}[t]
\centering
\includegraphics[width=0.46\textwidth]{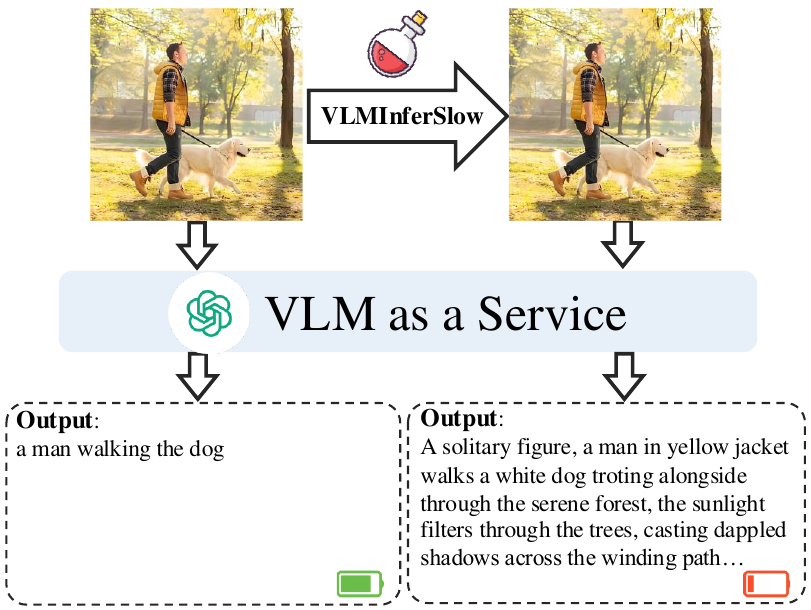}
\caption{Our \tool\ attack adds perturbations to input images, causing VLMs to generate longer sequences, resulting in reduced inference efficiency.}
\label{fig:topic}
\end{figure}

% \TODO{existing work and their limitation}

While prior work mainly focuses on optimizing the accuracy of VLMs, their robustness in terms of efficiency remains largely unexplored. Adversarial attacks are a widely used approach to evaluate the robustness of machine learning models.  
Although some adversarial attacks have targeted VLM inference efficiency, they all operate under an unrealistic assumption---namely, the white-box assumption. In real-world scenarios, however, VLMs are predominantly deployed as API services, making it unlikely for an attacker to have access to model parameters or architectures. Thus, existing methods may fail to accurately reflect the real-world threat of these models.
To bridge this gap, in this paper, we seek to answer the following question:

\begin{center}
\begin{tcolorbox}[colback=gray!10,%gray background
colframe=black,% black frame colour
width=7.7cm,% Use 8cm total width,
arc=1mm, auto outer arc,
boxrule=0.98pt,
]
\textit{Can we make unnoticeable adversarial inputs to significantly increase the computational consumption of VLMs with only the VLM inference API?}
\end{tcolorbox}
\end{center}

% \TODO{Our approach}
To address the aforementioned question,  we introduce \tool. 
Unlike existing works, evaluating the efficiency and robustness of VLMs using only their inference API presents several unique challenges. Firstly, without access to the model architecture and parameters, gradient-based approaches are unavailable.
To overcome this problem, we propose a novel zero-order optimization method, which relies solely on objective function values rather than gradients.
While the derivative-free optimization enables black-box evaluation, it introduces another challenge: zero-order methods may struggle when the objective function exhibits sharp changes in the loss surface. To mitigate this issue, we develop a fine-grained objective modeling approach tailored to our adversarial goals, increasing the victim VLM’s computational resource consumption. 
% \textcolor{red}{Lastly, VLMs are not static during inference (i.e., their output length varies), making existing ``static'' search strategies ineffective. To address this, we introduce a dynamic importance adjustment strategy that assigns different importance weights to each output token, enabling the adversarial search to focus on output tokens that most impact model efficiency.}

We evaluate \tool\ on four widely used VLMs across two datasets against four baselines. Experimental results demonstrate that \tool\  significantly increases the computational cost of VLMs up to 128.47\%, outperforming existing methods in the black-box setting significantly.   Moreover, comparisons with white-box baselines show that despite operating in a black-box setting, \tool\ achieves effectiveness comparable to white-box methods, which require access to the VLMs' architecture and parameters.  
Further experiments on adversarial examples quality, defense evaluation, robustness to different sampling strategies, and an ablation study validate the effectiveness and generalization ability of \tool.

% We evaluate our SAME using two widely-used
% multi-exit strategies (entropy-based (Xin et al.,
% 2020) and patience-based (Zhou et al., 2020)) with
% various pre-trained language models (Devlin et al.,
% 2019; Liu et al., 2019; Lan et al., 2020) as the backbone on eight tasks from the GLUE benchmark.
% Experimental results show that our SAME can effectively reduce the computational saving by 80%
% on average, which significantly outperforms previous accuracy-oriented approaches by a large margin. Further experiments on the multi-goal attack,
% attacking transferability, and adversarial training
% convincingly validate the effectiveness and generalization ability of our proposed SAME.

% However, considering the discrete efficiency property of VLMs, 

% \TODO{Contribution}

We summarize our contribution as follows:

\begin{itemize}
    \item \textbf{\textit{Problem Novelty:}} To the best of our knowledge, we are the first to study the efficiency robustness of VLMs under a black-box setting. This scenario better reflects real-world scenarios in which commercial VLMs are deployed as API services, providing a more accurate assessment of potential threats.

    \item \textbf{\textit{Technical Novelty:}} We design and implement \tool, which applies zero-order optimization and fine-grained \textit{efficiency modeling} with a dynamic importance strategy to assess the efficiency robustness of VLMs through adversarial attacks.

    \item \textbf{\textit{Empirical Evaluation:}} We conduct a systematic evaluation of various VLMs, and the results show that an adversary can generate imperceptible inputs that significantly increase the computational cost of VLMs up to 128.47\%. This highlights the need for future research on improving and safeguarding the efficiency robustness of VLMs.

\end{itemize}
% \clearpage

\section{Background \& Related Work}

% Vision-Language Models (VLMs)~\cite{alayrac2022flamingo,li2022blip,wang2022git,li2023blip,liu2024improved} typically adopt an encoder-decoder architecture. The encoder takes an image $\bm{x}$ and a text $\bm{s}_{in}$ as input and embeds them into the hidden representation, which is then sent to the decoder to generate the output sequence $\bm{y}=\{y_1, y_2, \cdots, y_N\}$. Starting with a special token such as beginning-of-sequence (BOS), the decoder generates sequences in a probabilistic and auto-regressive manner. More specifically, for the $i$-th generated token $y_i$, it is sampled over a probability distribution $f_i(\bm{x}, \bm{s}_{in}, y_1, y_2, \cdots, y_{i-1})\in\mathbb{R}^{V}$, where $V$ is the size of the  text vocabulary. 
% % For simplicity, we abbreviate it as $f_i(\bm{x})$ since we only make crafts on the image in this work. 
% The sequence generation process stops until the preset maximum length of the sequence is reached or the end-of-sequence (EOS) token is sampled, that is, $y_N=\text{EOS}$.

% Due to the auto regressive nature of the sequence generation process, the computational consumption is proportional to the number of decoder calls~\cite{chen2022nicgslowdown, gaoinducing}. As a result, longer generated sequences lead to increased computational consumption and reduced inference efficiency in VLMs.

% \subsection{Vision-Language Models}
\fakeparagraph{Vision-Language Models.}
\label{sec:background}
VLMs~\cite{li2022blip,alayrac2022flamingo,wang2022git,xiao2024florence} are a class of multimodal architectures designed to process image and text data simultaneously. 
% These models combine computer vision (CV) and natural language processing (NLP) techniques to establish correlations between the two modalities through shared embeddings. 
Typically, VLMs adopt an encoder-decoder architecture $\mathcal{F}(\cdot) = \{\mathcal{\mathcal{E}(\cdot), \mathcal{D}(\cdot)}\}$, where the input 
$\bm{x}=\{\mathcal{I}, \mathcal{T}\}$ 
consists of an image $\mathcal{I}$ and a text prompt $\mathcal{T}$. The encoder transforms $\mathcal{I}$ and $\mathcal{T}$ into hidden representations, which are integrated into a unified representation $\bm{h}$ and then fed into the decoder for generation.
% , denoted as $\bm{x}=\{\mathcal{I}, \mathcal{T}\}$.
% , with $\mathcal{I}$ refers to the input image and $\mathcal{T}$ representing the prompt text. 
% The encoder transforms the image and text prompt into hidden representations, which are then fused to form a unified representation. This fused representation is subsequently passed to the decoder for generation.
The decoding process in VLMs is autoregressive, starting with a special token (e.g., beginning-of-sequence or BOS token) and generating tokens sequentially. Formally, the decoder produces the $i$-th token by taking the representation $\bm{h}$
and preceding tokens as input.
% Formally, to produce the $i$-th
% \(i^{\text{th}}\) \textcolor{red}{$i$-th} 
% output token, the decoder takes the hidden representation \(\bm{h}\) and the previous tokens as input.
% \((i-1)^{th}\) \textcolor{red}{$(i-1)$-th} output token as input \textcolor{red}{only $y_{i-1}?$}. 
It computes the probabilities
% likelihood \textcolor{red}{probabilities} 
of the \(i\)-th token over the entire vocabulary \(\mathcal{V}\), denoted as $\text{Pr}(y_i|\, \mathcal{I})=\mathcal{D}(\bm{h};y_1,\cdots,y_{i-1})$, and then samples $y_i$ on \(\mathcal{V}\) based on this distribution.
% \(\text{Pr}(y_i) = \mathcal{D}(\bm{h}, y_{i-1})\)
The autoregressive nature of this process inherently prevents parallel token generation, as each token depends on previously generated tokens. Longer sequences therefore lead to more decoder calls, and reduced inference efficiency, with both computational cost and inference time growing in proportion to the length of the generated sequence.

% Due to the autoregressive nature of this process, token generation cannot occur in parallel. Consequently, generating longer sequences requires additional computational resources, as each token depends on all previously generated tokens. This results in increased decoder calls and reduced inference efficiency, with both computational cost and inference time growing in proportion to the length of the generated sequence.

% \subsection{DNNs Efficiency}
\fakeparagraph{DNNs Efficiency.}
High model accuracy typically entails a large model, leading to substantial computational costs and low efficiency. 
 % DNNs' timely response is an important aspect of real-world applications.
% For instance, in systems designed to transfer visual information into audio for people with visual impairments~\cite{mackowski2023multimodal,masal2024development}, slow response times can degrade the user experience and hinder practical usability. 
Endeavors have been made towards fastening the inference process. Existing works include offline pruning redundant neurons~\cite{kurtz2020inducing,hoefler2021sparsity}, and adaptively skipping some parts during inference~\cite{zhou2020bert,meng2022adavit} to reduce computational consumption. However, these methods are not robust against adversarial attacks~\cite{haque2022ereba,haque2023antinode,zhang2023slowbert,chen2023dynamic}, i.e., they cannot effectively reduce computational consumption when processing adversarial inputs. 

% Malicious attackers can send adversarial inputs to the model to reduce DNNs' efficiency. 

\begin{figure}[t]
\centering
\includegraphics[width=0.38\textwidth]{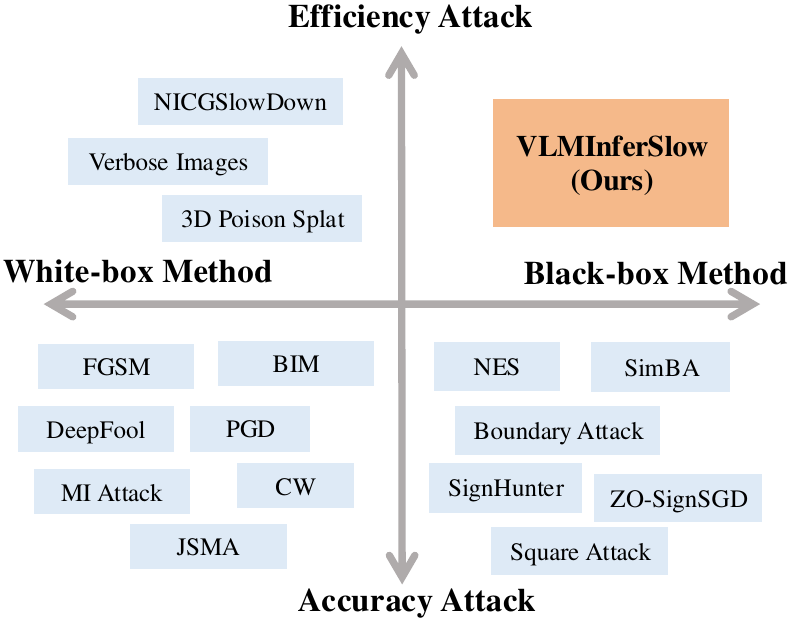}
\caption{Comparison of \tool\ and existing works in terms of attack goals (accuracy vs. efficiency) and attack types (white-box vs. black-box).}
\label{fig:category}
\end{figure}

% \subsection{Adversarial Attacks}
\fakeparagraph{Adversarial Attacks.}
Adversarial attacks aim to fool the model by modifying benign input. Most works target the accuracy surface, aiming to reduce the accuracy of victim models. Based on the accessibility of the full model, they can be categorized into white-box methods~\cite{shayegani2023jailbreak,qi2024visual,zhang2024adversarial,chang2024play} and black-box methods~\cite{guo2019simple,al2020sign,zhao2023evaluating,cheng2024efficient}.
% white-box methods~\cite{goodfellow2014explaining,fredrikson2015model,kurakin2016adversarial,moosavi2016deepfool,papernot2016limitations,madry2017towards,carlini2017towards,shayegani2023jailbreak,qi2024visual,zhang2025adversarial} and black-box methods~\cite{ilyas2018black,liu2019signsgd,guo2019simple,brendel2017decision,al2020sign,andriushchenko2020square,cheng2024efficient}.

Another important but overlooked problem is the efficiency vulnerability of DNNs. A few studies have explored the efficiency attack~\cite{feng2024llmeffichecker,chen2023dark,chen2022nmtsloth,haque2023antinode}. For example, NICGSlowdown~\cite{chen2022nicgslowdown} delays the occurrence of EOS token to increase decoder calls in the image captioning task, while Verbose Images~\cite{gaoinducing} designs specific losses for VLMs to increase computational consumption. However, these methods are confined to the white-box attack paradigm, while 
% limiting their practicality in real-world scenarios. In contrast, 
our work pioneers the investigation of efficiency attack in a more practical black-box setting.
Fig.~\ref{fig:category} illustrates the difference between our approach and existing methods. 
% To the best of our knowledge, this is the first study to investigate the efficiency vulnerability of VLMs in a black-box scenario.

\section{Preliminary}

\subsection{Threat Model}
\label{sec:threat}
\fakeparagraph{Adversarial Goal.}
Unlike existing adversarial attacks that primarily aim to compromise a model's \textit{integrity}, our attack specifically targets the \textit{availability} of VLMs, with the goal of disrupting their functionality and accessibility. The aim is to create imperceptible adversarial images that significantly increase the computational resource consumption of target models. This is achieved by forcing VLMs to generate excessively long sequences. Moreover, the adversarial images should be indistinguishable from benign images to human observers while maintaining realism in real-world contexts.

Such an attack could have severe consequences. For example, if VLMs are deployed on mobile devices, our attack could drain the device’s battery, rendering it unusable. Similarly, if VLMs are deployed on servers offering services, the attack could occupy GPU memory, degrade service quality, and cause disruptions for legitimate users.

\fakeparagraph{Adversarial Assumption and Capability.} 
% Given a victim VLM that uses an autoregressive process to generate outputs (as described in Sec.~\ref{sec:background}), 
Our approach contrasts with existing works~\cite{gaoinducing} that assume full access to the VLM's architecture and parameters, which is unrealistic. 
% We do not make such unrealistic assumptions. 
Instead, we consider a more realistic machine learning as a service (MLaaS) scenario, where the adversary behaves as a benign user, querying the VLM's API with inputs. In this scenario, the deployed VLM API returns the corresponding textual outputs and logits. 
This scenario is valid and realistic, as most mainstream commercial VLM providers, including OpenAI, Google Gemini, and others, deploy their model APIs in this manner.
% \textcolor{red}{does it overclaim? nearly all? does not highlight that probability distribution is accessible.}

\subsection{Problem Formulation}
\label{sec:problemformulation}
We consider a victim VLM \(\mathcal{F}(\cdot)\) with input \(\bm{x} = (\mathcal{I}, \mathcal{T})\), where \(\mathcal{I}\) is input image and \(\mathcal{T}\) is text prompt. Our work focuses on the image modality of the VLM's input, as some commercial VLMs do not allow modification of the hidden input prompt \(\mathcal{T}\). The goal of our attack is to find an optimal image \(\mathcal{I}^\prime\) that satisfies the conditions described in Sec.~\ref{sec:threat}.

As stated in Sec.~\ref{sec:threat}, the adversarial goal is to generate human-unnoticeable perturbations to images to decrease the victim VLMs' efficiency during inference. Specifically, the adversarial objective concentrates on three factors: \textbf{(1) \textit{Effectiveness}}. The generated adversarial image should increase the victim VLM's computational resource consumption; \textbf{(2) \textit{Unnoticeability}}. The adversarial image cannot
be differentiated by humans from the benign image; and \textbf{(3) \textit{Realistic}}. The adversarial image should be realistic in the real world.
\begin{equation}
\label{eq:define}
    \begin{split}
        & \quad \Delta = \text{argmax}_{\delta} \ \mathcal{RC}_{\mathcal{F}}(\mathcal{I} + \delta) \\ 
        & s.t. 
        % \quad \quad \quad 
        \ 
 \|\delta\| \le \epsilon \; \wedge \; (\mathcal{I}+\delta) \in [0, 1]^n \\
    \end{split}
\end{equation}

We formulate our problem as a constrained optimization problem in Eq.~\ref{eq:define}, where \(\mathcal{I}\) is the benign input, \(\mathcal{F}\) denotes the victim VLM under attack, \(\epsilon\) is the maximum allowable adversarial perturbation, and \(\mathcal{RC}_{\mathcal{F}}(\cdot)\) measures the resource consumption of \(\mathcal{F}\) when processing a given input. 
Our proposed approach, \tool, aims to identify an optimal perturbation \(\Delta\) that maximizes the computational resources required to handle \(\mathcal{I} + \delta\), while ensuring the perturbation remains imperceptible to humans (i.e., $\|\delta\| \le \epsilon$) and preserves realism in real-world scenarios (i.e., $(\mathcal{I}+\delta) \in [0, 1]^n$, where $n$ denotes the dimension of the image input).

\section{Approach}

\subsection{Design Overview}

\begin{figure}[t]
\centering
\includegraphics[width=0.475\textwidth]{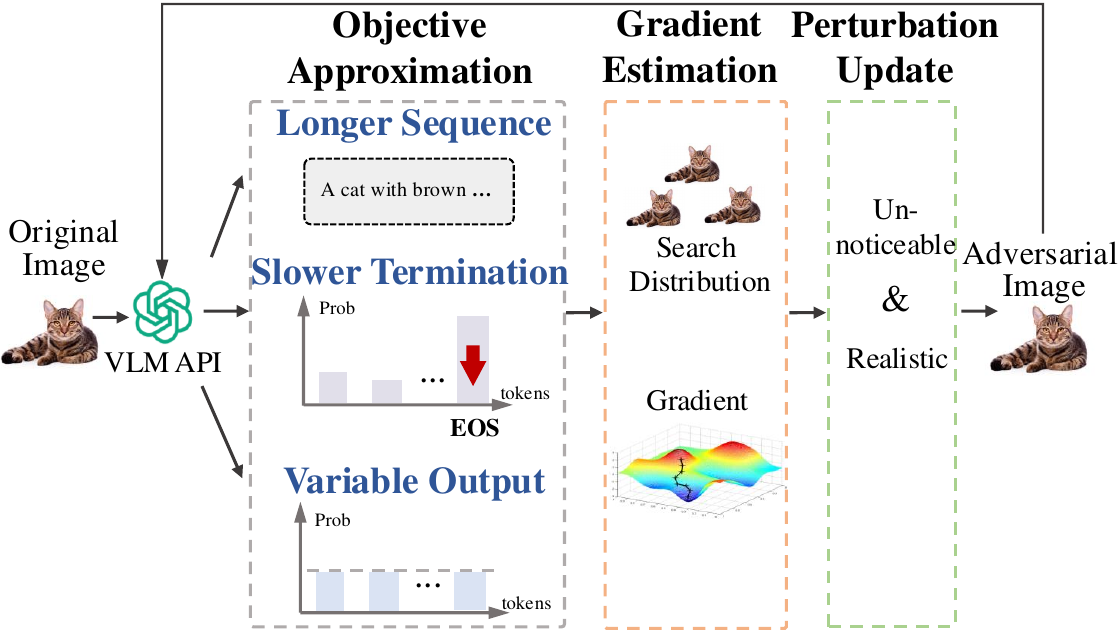}
\caption{Design overview of \tool.}
\label{fig:methodoverview}
\end{figure}

The framework of our method \tool\ is illustrated in Fig~\ref{fig:methodoverview}. We iteratively modify the input data to generate adversarial images. For each iteration, firstly, we design objectives to approximate the adversarial goal (Sec.~\ref{sec:objectapproximation}). Secondly, without access to model architectures and parameters, we propose a zero-order optimization module to estimate the gradient (Sec.~\ref{sec:gradientestimation}). After this, we update the adversarial image while satisfying the perturbation constraint
(Sec.~\ref{sec:update}).

\subsection{Adversarial Objective Approximation}
\label{sec:objectapproximation}
Our attack aims to maximize the resource consumption $\mathcal{RC}_{\mathcal{F}}(\cdot)$ (Eq.~\ref{eq:define}). However, there is no existing objective to represent this metric. As stated in Sec.\ref{sec:background}, longer sequences lead to more decoder calls, and thus reduce VLMs' efficiency. Motivated by this, we approximate the $\mathcal{RC}_{\mathcal{F}}(\cdot)$ by designing three efficiency-oriented adversarial objectives, elaborated as follows.

\fakeparagraph{Longer Sequence Generation.} The most straightforward target is to prolong the length of the output sentence to increase the number of decoder calls. It is notable that even if this objective is non-differentiable, it suits our derivative-free approach since no actual gradient is calculated.
\begin{equation}
\mathcal{L}_{len}(\delta) =\textbf{Length}(\mathcal{F}(\mathcal{I}+\delta)). 
\end{equation}
% \TODO{$\mathcal{L}_{len}(x) =\text{Length}(\mathcal{F}(x))$}

\fakeparagraph{Slower Termination Occurrence.} We extend the generated sequence by slowing the occurrence of the termination signal, i.e., the end-of-sequence (EOS) token. Denote the probability of the $i$-th token as $\text{Pr}(y_i| \, \mathcal{I}+\delta)$, which is assumed to be accessible. We decrease the corresponding probability of the EOS token. Moreover, VLMs are not static during inference (i.e., their output length varies). Considering this, we introduce a dynamic weight decay strategy to enable the adversarial search to focus on output tokens that most impact model efficiency. Specifically, greater weights are attached to probabilities whose positions are closer to the end of the sequence. 
Formally, it is:
\begin{equation}
\mathcal{L}_{eos}(\delta) =  -\sum_{i=1}^{N}\omega^{N-i}\text{Pr}^{\text{EOS}}(y_i \, | \, \mathcal{I}+\delta),
\end{equation}
where $\text{Pr}^{\text{EOS}}(y_i\, | \, \mathcal{I}+\delta)$ is the probability of the $i$-th output token in the sequence being sampled as an EOS token, $N$ is the length of the generated sequence, and $\omega<1$ is the hyper-parameter for controlling the weight decay speed. We set $\omega=0.1$ in practice. 

% \TODO{ what is $\text{Pr}^{\text{EOS}}(y_i)$, pleasz explain each symbol in each equation, same for following equation, and each L is a funciton, it should includes a input}

\fakeparagraph{Variable Output Production.} We propose to encourage VLMs to produce more variable and less predictable tokens, thereby resulting in more complex and longer sequences. We achieve this by aligning the probability distribution $\text{Pr}(\cdot)$ of each token more closely with a discrete uniform distribution $\mathcal{U}$. In this way, the probability of diverse token candidates being sampled is increased. However, the vocabulary size is usually huge, making the objective difficult to optimize. To tackle this, for each token, we extract the top-$k$ probabilities and normalize them to form a new probability distribution $\Tilde{\text{Pr}}(\cdot)$. The objective is then formulated as the sum of the KL divergence between $\Tilde{\text{Pr}}(\cdot)$ and $\mathcal{U}$ across all positions:
\begin{equation}
    \mathcal{L}_{var}(\delta) =-\frac{1}{N}\sum_{i=1}^{N}D_{\text{KL}}(\Tilde{\text{Pr}}(y_i\, | \, \mathcal{I}+\delta) \, \Vert \, \mathcal{U}).
\end{equation}
In practice, we set $k=100$. Tokens at all positions are assigned equal weights, as they collectively enhance the variability of the generated sentence.

\fakeparagraph{Final Objective.} Our final objective is:
\begin{equation}
    \mathcal{L}(\delta) = \mathcal{L}_{len}(\delta) + \alpha\mathcal{L}_{eos}(\delta) + \beta\mathcal{L}_{var}(\delta),
\label{eq:totol_loss}
\end{equation}
where $\alpha$ and $\beta$ are hyper-parameters for weighting different objectives.

\subsection{Gradient Estimation}
\label{sec:gradientestimation}
As stated in Eq.~\ref{eq:define}, we optimize the perturbation $\delta$ by maximizing the approximated adversarial objective $\mathcal{L}(\delta)$ for each input image $\mathcal{I}$. Since the gradient-based approach is unavailable in our black-box setting, we adopt a derivative-free optimization method to estimate the gradient.
% We optimize the perturbation $\delta$ by maximizing the loss $\mathcal{L}$. However, in the black-box setting, the gradient  $\nabla_{\delta}\mathcal{L}(\delta)$ is not directly obtainable. Therefore, we employ a zeroth-order optimization algorithm to estimate the gradient. 
% Motivated by
Following Natural Evolution Strategies~\cite{wierstra2014natural,ilyas2018black}, we maximize the expected value of the objective under a search distribution $\pi(z|\delta)$: 
\begin{equation}
    J(\delta) = \mathbb{E}_{\pi(z|\delta)}[\mathcal{L}(z)]=\int \mathcal{L}(z)\pi(z|\delta)\ dz.
\end{equation}
Then, the gradient of $J(\delta)$ is computed as:
\begin{equation}
\nabla_{\delta}J(\delta)=
\mathbb{E}_{\pi(z|\delta)}[\mathcal{L}(z)\nabla_{\delta}\log\pi(z|\delta)].
\end{equation}
The detailed derivation is provided in Appendix~\ref{sec:appendix_derivation}. We take the search distribution $\pi(z|\delta)$ as $\mathcal{N}(\delta,\eta^2I)$, where $\eta$ is search variance. Following~\citet{salimans2017evolution}, we sample a population of $z_i$ as follows. Firstly, we sample $q$ gaussian noises $\mu_j$ from $\mathcal{N}(0,I)$, and set $\mu_{q+j}=-\mu_j$, $j\in\{1,2,\cdots,q\}$. Then, we use $z_i=\delta+\eta\mu_i$ to obtain $z_i$, $i\in\{1,2,\cdots,2q\}$. In this way, the gradient $\nabla_{\delta}J(\delta)$ is estimated as:
\begin{equation}
\hat{\nabla}_{\delta}J(\delta)=
% \approx
\frac{1}{2\eta q}\sum_{i=1}^{2q}\mu_i\mathcal{L}(\delta+\eta\mu_i).
\label{eq:gradient}
\end{equation}

This estimation is theoretically guaranteed:
\begin{theorem}[\citealp{ilyas2018black}]
Denote $\hat{\nabla}$ as the estimation of gradient and $\nabla$ as the true gradient. As search variance $\eta\rightarrow0$, we have:
\[
\resizebox{1.0\linewidth}{!}{$
\mathbb{P}\left\{(1-\zeta)\|\nabla\|^2 \leq \|\hat{\nabla}\|^2 \leq (1+\zeta)\|\nabla\|^2\right\} \geq 1 - 2p
$,}
\] where $0<\zeta<1$ and $q=
\mathcal{O}(-\zeta^{-2}\log(p))$.
\end{theorem}

\subsection{Adversarial Example Update}
\label{sec:update}
After obtaining the estimation of gradient $\hat{\nabla}_{\delta}J(\delta)$, we update the perturbation via gradient ascent while adhering to the constraint. The perturbation is updated as $\delta=\delta+\gamma\times\hat{\nabla}_{\delta}J(\delta)$
% /\|\nabla_{\delta}J(\delta)\|$
, optimizing toward maximizing the approximated objective (Eq.~\ref{eq:totol_loss}). Then, we clip the updated perturbation to adhere to the \textit{unnoticeability} constraint $\|\delta\|\le\epsilon$ (Eq.~\ref{eq:define}). Formally, it is:
\begin{equation}
  \text{Clip}(\delta,\epsilon) = 
\begin{cases} 
\delta & \text{if $\|\delta\|\le\epsilon$}; \\
\epsilon\times \frac{\delta} {\|\delta\|} & \text{else}.
\end{cases}  
\label{eq:clip}
\end{equation}
This constraint limits the $L_2$ norm of perturbation added to the image $\mathcal{I}$ to a maximum of $\epsilon$.
After resizing the perturbation \(\delta\) to ensure it satisfies the imperceptibility constraints, we also apply a clipping operation to the perturbed image to ensure it meets the \textit{realistic} constraint, consistent with existing work~\cite{madry2017towards}.
The complete procedure of our \tool\ attack is summarized in the Algorithm~\ref{algorithm} in Appendix~\ref{sec:algorithm}.

\section{Evaluation}
\label{sec:evaluation}

\begin{table*}[t]
  \centering
  \small
   \resizebox{0.88\textwidth}{!}{
    \begin{tabular}{cccccccc}
    \toprule
    \multirow{2}[2]{*}{\textbf{Models}} & \multirow{2}[2]{*}{\textbf{Methods}} & \multicolumn{3}{c}{\textbf{MS-COCO}} & \multicolumn{3}{c}{\textbf{ImageNet-1k}} \\
          &       & \textbf{I-length} & \textbf{I-latency} & \textbf{I-energy} & \textbf{I-length} & \textbf{I-latency} & \textbf{I-energy} \\
    \midrule
    \multirow{6}[2]{*}{Flamingo} & Gaussian & -4.15 & -0.16 & -6.92 & -4.27 & -1.12 & 6.96 \\
          % & TVM   & 0.51  & 5.55  & 11.27 & 1.25  & 0.18  & 8.92 \\
          & JPEG  & -7.92 & -0.13 & -5.03 & -3.95 & -5.26 & 6.28 \\
          & NICGSlowdown-B & -3.54 & 0.19  & -0.22 & -1.14 & -0.12 & -1.74 \\
          & Verbose-B & -2.93 & 5.56  & -0.13 & -0.63 & 5.26  & -1.70 \\
          & \tool & \textbf{128.47} & \textbf{105.56} & \textbf{115.19} & \textbf{103.44} & \textbf{78.42} & \textbf{70.32} \\
    \midrule
    \multirow{6}[1]{*}{BLIP} & Gaussian & 18.92 & 18.19 & 26.43 & 20.50  & 20.42 & 24.50 \\
          % & TVM   & 16.24 & 9.09  & 25.17 & 19.39 & 21.34 & 24.99 \\
          & JPEG  & 28.86 & 27.27 & 39.34 & 35.87 & 42.43 & 37.54 \\
          & NICGSlowdown-B & -9.40  & -9.09 & -3.54 & 6.09  & 11.24 & 4.29 \\
          & Verbose-B & -6.98 & 4.24  & -0.84 & 9.02  & 15.28 & 7.06 \\
          & \tool& \textbf{71.95} & \textbf{54.98} & \textbf{65.41} & \textbf{74.38} & \textbf{66.89} & \textbf{55.58} \\
    \midrule
    \multirow{6}[0]{*}{GIT} & Gaussian &-18.33  &-5.21  & -16.68 &42.56  &38.46   & 15.54 \\
          % & TVM   &9.29  &1.01  & 1.94 &4.67  &6.15   &9.43  \\
          & JPEG & 21.47 & 30.09 & 17.25 & 14.53 & 13.85  & 5.67 \\
          & NICGSlowdown-B & 4.27 & 6.67 & 5.89 & 11.64 & 16.15  &17.79  \\
          & Verbose-B & 8.23 & 9.31 & 11.22 &18.21  & 8.48  &9.21  \\
          & \tool & \textbf{75.93} & \textbf{66.67} & \textbf{78.59} & \textbf{93.86} & \textbf{115.38} & \textbf{84.82} \\
    \midrule
    \multirow{6}[1]{*}{Florence} & Gaussian & 1.75  & 3.03  & 0.61  & -0.40  & -3.57 & -1.92 \\
          % & TVM   & -8.50  & -9.09 & -10.78 & -2.18 & -5.36 & -2.61 \\
          & JPEG  & 0.02  & 3.03  & -1.27 & -0.83 & -5.36 & -0.28 \\
          & NICGSlowdown-B & -2.56 & -6.59 & -4.54 & -2.07 & -1.79 & 0.40 \\
          & Verbose-B & -3.01 & -6.06 & -7.88 & -3.08 & -1.78 & -2.51 \\
          &\tool &\textbf{51.96} & \textbf{42.42} & \textbf{51.50} & \textbf{47.39} & \textbf{42.86} & \textbf{65.83} \\
    \bottomrule
    \end{tabular}%
    }
\caption{Results of the relative increase in sequence length (I-length, \%), in response latency (I-latency, \%), and in energy consumption (I-energy, \%). NICGSlowdown-B and Verbose-B refer to that we evaluate these two white-box methods in the black-box setting. Best results are in \textbf{bold}.}
\label{tab:maintable}%
\end{table*}%

\subsection{Evaluation Setup}

\fakeparagraph{Datasets and Models.} 
We evaluate our approach on the image captioning task using two public datasets containing images from diverse scenes: MS-COCO~\cite{lin2014microsoft} and ImageNet~\cite{deng2009imagenet}. 
As for the victim models, we choose four different vision-language models, which are \textsc{Flamingo}~\cite{alayrac2022flamingo}, \textsc{Blip}~\cite{li2022blip}, \textsc{Git}~\cite{wang2022git} and \textsc{Florence}~\cite{xiao2024florence}. 
% These VLMs have distinct structures and generate tokens in an autoregressive manner. 
We use their default text prompt for our task.  More details of these models are provided in Appendix~\ref{sec:appendix_modelanddetails}.

% \TODO{introduce each model with one to two sentences}

% \TODO{introduce each datasets with few sentences}

\fakeparagraph{Comparison Baselines.} To the best of our knowledge, we are the first to evaluate the efficiency and robustness of VLMs in the black-box setting, with no existing off-the-shelf black-box baselines. To address this, we compare \tool\ against (1) two state-of-the-art \textit{white-box} approaches, NICGSlowdown~\cite{chen2022nicgslowdown} and Verbose Images~\cite{gaoinducing}; and (2) two widely-used \textit{natural image corruptions}, including Gaussian noise~\cite{xu2017feature,hendrycks2019benchmarking}
% feature squeezing (TVM)~\cite{xu2017feature}, 
% image quantization~\cite{xu2017feature,hendrycks2019benchmarking},
and JPEG compression~\cite{liu2019feature}, as our comparison baselines. 

\fakeparagraph{Black-box Evaluation Setting.}
For \textit{black-box} methods (Gaussian, JPEG, and \tool), adversarial images are directly optimized and evaluated on the target model itself. For \textit{white-box} methods (NICGSlowdown and Verbose), they do not support the black-box setting. Therefore, following the existing transferability setting, adversarial images are generated using accessible surrogate models (excluding the target model) and then transferred to the target VLM for evaluation.

% Specifically, we choose two white-box methods NICGSlowdown~\cite{chen2022nicgslowdown} and Verbose Images~\cite{gaoinducing}, and three image corruptions including gaussian noise~\cite{xu2017feature,hendrycks2019benchmarking}, feature squeezing (TVM)~\cite{xu2017feature}, 
% % image quantization~\cite{xu2017feature,hendrycks2019benchmarking},
% and JPEG compression~\cite{liu2019feature} as our comparison baselines.

% \TODO{introduce more for each methods}

\fakeparagraph{Evaluation Metrics.}
Following~\citet{chen2022nicgslowdown} and~\citet{gaoinducing}, we use three metrics: the relative increase in sequence length (I-length), in response latency (I-latency), and in energy consumption (I-energy), to represent the inference efficiency of VLMs. Their formal definitions are:
\begin{equation*}
\begin{aligned}
    \text{I-length} &=\frac{\text{length}(\mathcal{I}+\delta) - \text{length}(\mathcal{I})}{\text{length}(\mathcal{I})}  \times 100\%  \\
    \text{I-latency} &=\frac{\text{latency}(\mathcal{I}+\delta) - \text{latency}(\mathcal{I})}{\text{latency}(\mathcal{I})}\times 100\%   \\
    \text{I-energy} &= \frac{\text{energy}(\mathcal{I}+\delta) - \text{energy}(\mathcal{I})}{\text{energy}(\mathcal{I})}\times 100\%,
\end{aligned}
\end{equation*}
where $\mathcal{I}$ is the original image and $\delta$ is the added perturbation. $\text{length}(\cdot)$,  $\text{latency}(\cdot)$, $\text{energy}(\cdot)$ are functions to calculate the sequence length, response latency, and energy consumption respectively.

\begin{table}[tbh!]
  \centering
\resizebox{0.475\textwidth}{!}{
    \begin{tabular}{ccccc}
    \toprule
    \textbf{Models} & \textbf{Methods} & \textbf{I-length} & \textbf{I-latency} & \textbf{I-energy} \\
    \midrule
    \multirow{3}[1]{*}{Flamingo} & NICGSlowdown &     47.62 & 44.44 & 49.08  \\
          & Verbose &  122.39 & 105.32 & 97.89  \\
          & \tool &      \textbf{128.47} & \textbf{105.56} & \textbf{115.19}  \\
          \midrule
    \multirow{3}[0]{*}{BLIP} & NICGSlowdown &   53.83 & 44.55 & 53.21 \\
          & Verbose &   \textbf{125.10} & \textbf{90.21} & \textbf{96.43} \\
          & \tool &    71.95 & 54.98 & 65.41 \\          \midrule
    \multirow{3}[0]{*}{GIT} & NICGSlowdown &44.46  & 59.62 & 43.10 \\
          & Verbose & \textbf{100.31} &\textbf{106.93}  &\textbf{95.21} \\
          & \tool &     75.93 & {66.67} & {78.59}  \\          \midrule
    \multirow{3}[1]{*}{Florence} & NICGSlowdown &      43.05 & 36.97 & 45.47 \\
          & Verbose &  20.31 & 18.18 & 31.09  \\
          & \tool &  \textbf{51.96} & \textbf{42.42} & \textbf{51.50}  \\
    \bottomrule
    \end{tabular}%
}
\caption{Comparison with two white-box baselines on MS-COCO. Best results are in \textbf{bold}. Results for ImageNet-1k are in Appendix~\ref{sec:appendix_addition}.}
\label{tab:whitebox}
\end{table}%

\fakeparagraph{Implementation Details.} For each model, we update the perturbation for $T=500$ iterations with step size $\gamma=5$.  We set $\alpha=0.5$ and $\beta=0.1$ in the objective function (Eq.~\ref{eq:totol_loss}).  More details are provided in Appendix~\ref{sec:appendix_modelanddetails}.

% \subsection{Effectiveness and Severity}

% \fakeparagraph{Metrics and Process}

% \fakeparagraph{Results and Analysis}

\subsection{Main Results}
\label{sec:mainresults}
\fakeparagraph{Results in Black-box Setting.}
To evaluate the effectiveness and severity of our \tool\ attack in the black-box setting, we measure the I-length, I-latency, and I-energy against the four VLMs during inference. The results are present in Tab.~\ref{tab:maintable}. 
It demonstrates that compared to baseline methods, our \tool\ significantly increases three metrics on all four VLMs. Specifically, \tool\ achieves an average increase in I-length of 82.08\% and 76.98\%  on MS-COCO and ImageNet-1k, respectively. The baseline methods are ineffective in consistently performing well in the black-box setting, and some even cause a reduction in the metrics. This suggests that \tool\ is effective while simple natural image corruptions and white-box methods are not reliably applicable in practical black-box scenarios.

\begin{figure*}[t]
\centering
\includegraphics[scale=0.58]{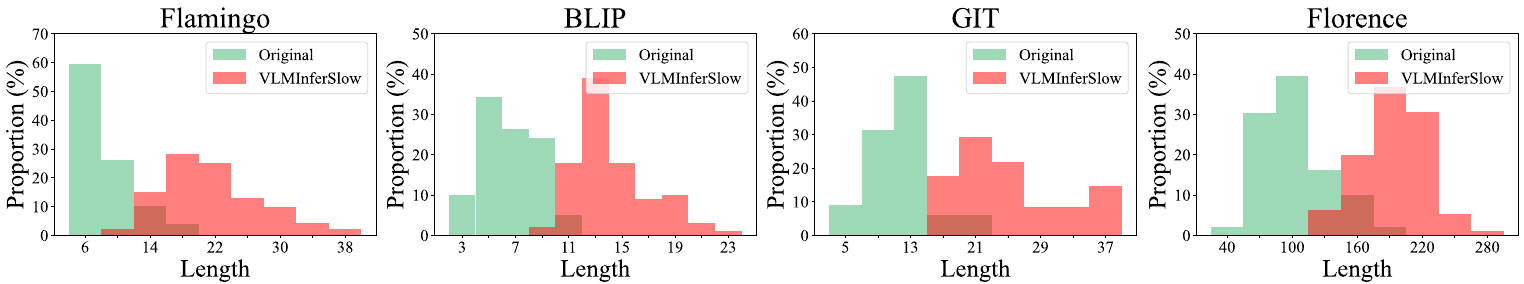}
\caption{The generated sequence 
 length distribution of four VLMs.}
\label{fig:distribution}
\end{figure*}

We present the distribution of the length of generated sequences on original images and our \tool\ generated adversarial images in Fig.~\ref{fig:distribution}. It shows that the sequences generated on our adversarial images tend to be longer than those on original images. Notably, our \tool\ works on diverse VLMs and settings. For example, \textsc{Flamingo} typically generates concise sequences, while \textsc{Florence} already produces long sequences. The \tool\ generated images can yield four different VLMs to generate longer sequences, demonstrating its effectiveness in reducing the inference efficiency of diverse VLMs.

\fakeparagraph{Comparison with White-box Baselines.}
We also compare \tool\ with two white-box methods under the white-box setting, where full access to the model is assumed. As shown in Tab.~\ref{tab:whitebox}, \tool, even in the black-box setting, achieves performance comparable to NICGSlowdown and Verbose Images in the white-box setting.

% In summary, \tool\ substantially increases the computation consumption of VLMs, surpassing all baselines. Notably, it achieves performance on par with white-box methods.

\begin{figure}[tbp]
\centering
\includegraphics[width=0.46\textwidth]{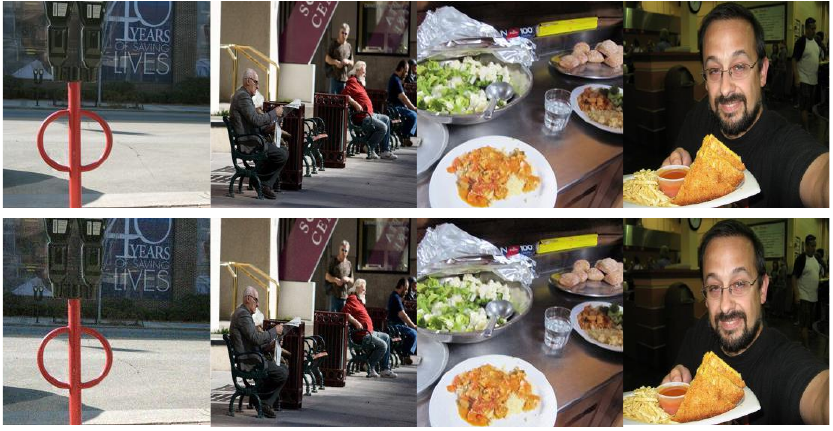}
\caption{Examples of original images (upper row) and adversarial images (lower row).}
\label{fig:image_example}
\end{figure}

\subsection{Quality of Generated Images}
\label{sec:qualityofgeneratedimages}

% \fakeparagraph{Metrics and Process}

% \TODO{which model for computing embedding distance}

We measure the $L_2$ distance and the image feature dissimilarity (detailed in Appendix~\ref{sec:appendix_modelanddetails}) between the original images and adversarial images generated by different methods. We use the image encoder of CLIP~\cite{radford2021learning} to extract the features of images.
The results, as shown in Tab.~\ref{tab:distance}, indicate that the average $L_2$ distance and image feature dissimilarity for adversarial images generated by \tool\ are 10.63 and 0.03, respectively. These values are slightly higher than those of the two white-box methods but much smaller than those of natural image corruptions. This is because the optimization of \tool\ is not as precise as white-box methods, leading to slightly larger $L_2$ distance and image feature dissimilarity scores than white-box methods. As for natural image corruptions, they simply modify or add noise to the original images, leading to obvious discrepancies between the original and generated images. Examples of benign images and their corresponding adversarial images are shown in Fig.~\ref{fig:image_example}, demonstrating that the perturbations are imperceptible to human observers. We also present examples of benign and adversarial images and their generated sequences in Appendix~\ref{sec:appendix_visualization}.

\begin{table}[htbp]
  \centering
\resizebox{0.46\textwidth}{!}{
    \begin{tabular}{ccccc}
    \toprule
    \textbf{Distances} & \textbf{Methods} & \textbf{COCO} & \textbf{IN-1k} & \textbf{Avg.} \\
    \midrule
    \multirow{5}[2]{*}{$L_2$} & Gaussian & 39.36      &40.51       &  39.94\\
          & JPEG  &   123.49    &   107.07    & 115.28 \\
          & NICGSlowdown &   3.40     &   3.77   & 3.59 \\
          & Verbose &  6.39     & 7.82      & 7.11 \\
          & \tool &   10.12    &  11.14     & 10.63 \\
    \midrule
    \multirow{5}[2]{*}{\makecell{Image \\ feature \\ dissimilarity}} & Gaussian &    0.20   &     0.19  & 0.20 \\
          & JPEG  &  0.43     & 0.43      & 0.43 \\
          & NICGSlowdown &  0.01     & 0.01     &0.01  \\
          & Verbose &  0.01     &  0.02     & 0.02 \\
          & \tool &   0.03    &   0.03    & 0.03 \\
    \bottomrule
    \end{tabular}%
    }
    \caption{$L_2$ and image feature dissimilarity of original and adversarial images. Results are for \textsc{Flamingo}.}
  \label{tab:distance}%
\end{table}%

\subsection{Ablation Study}
We study the effect of the approximated objectives on decreasing efficiency of VLMs. In Sec.~\ref{sec:objectapproximation}, we propose $\mathcal{L}_{len}$, $\mathcal{L}_{eos}$, and $\mathcal{L}_{var}$ as the optimization objectives, aiming to generate longer sequences, delay the generation termination, and enhance the variability of output. We conduct an ablation study by adopting one or two of the three objectives. The results are shown in Tab.~\ref{tab:ablation_loss}. It can be observed that each objective contributes to the reduced efficiency. More specifically, $\mathcal{L}_{len}$ has the most significant individual impact (e.g., 63.74\% for I-length on MS-COCO). Employing all three objectives yields the best performance, validating that each objective contributes to the reduced inference efficiency. We provide more ablation studies in Appendix~\ref{sec:appendix_addition}.

\begin{table}[tbph]
  \centering
  \small
  \resizebox{0.46\textwidth}{!}{
    \begin{tabular}{ccc|ccc}
    \toprule
    $\mathcal{L}_{len}$ & $\mathcal{L}_{eos}$ & $\mathcal{L}_{var}$ & \textbf{I-length} & \textbf{I-latency} & \textbf{I-energy} \\
    \midrule
    \checkmark     &       &       & 63.74 & 43.52 & 52.62 \\
          & \checkmark     &       & 33.02 & 20.13 & 26.31 \\
          &       & \checkmark     & 26.73 & 15.81 & 23.02 \\
    \checkmark     & \checkmark     &       &  93.73  &  71.69 &  72.36 \\
    \checkmark     &       & \checkmark     & 76.13 & 49.48 & 59.20 \\
          & \checkmark     & \checkmark     & 35.25 & 29.85 & 29.60 \\
    \checkmark     & \checkmark     & \checkmark    & \textbf{128.47} & \textbf{105.56} & \textbf{115.19} \\
    \bottomrule
    \end{tabular}%
    }
\caption{Ablation results of three designed objectives. Results are from \textsc{Flamingo} on MS-COCO.}
\label{tab:ablation_loss}
\end{table}

\section{Discussion} 
\label{sec:morestudies}

% Other discussions, such as robustness to different sampling strategies, are presented in Sec. \ref{sec:appendix_addition}.

% \TODO{different hyper-paramenter}

\fakeparagraph{Efficiency VS. Accuracy.} \tool\ is proposed to reduce the inference efficiency of VLMs. To further investigate whether it affects accuracy, we measured the BLEU score and text feature dissimilarity (detailed in Appendix~\ref{sec:appendix_modelanddetails}) of captions generated on original and adversarial images, using MS-COCO captions as references. The text features are extracted by BERT~\cite{devlin2018bert}. Results are presented in Tab.~\ref{tab:bleu}. It can be observed that \tool\ reduces the BLEU score by up to 38.46\% while increasing the text feature dissimilarity by up to 14.81\%. This demonstrates that \tool\ not only significantly reduces the efficiency, but also lowers the accuracy of VLMs.

\begin{table}[htbp]
  \centering
  \small
\resizebox{0.46\textwidth}{!}{
    \begin{tabular}{ccccc}
    \toprule
    \textbf{Metrics} & \textbf{Models} & \textbf{Ori.} & \textbf{Adv.} & \textbf{Change (\%)} \\
    \midrule
    \multirow{4}[2]{*}{BLEU} & Flamingo & 0.21  & 0.15  & 28.57 ($\downarrow$)  \\
          & BLIP  & 0.24  & 0.19  & 20.83 ($\downarrow$)\\
          & GIT   & 0.30   & 0.20   & 33.33 ($\downarrow$) \\
          & Florence & 0.13  & 0.08  & 38.46 ($\downarrow$) \\
    \midrule
    \multirow{4}[2]{*}{\makecell{Text \\ feature \\ dissimilarity}} & Flamingo & 0.29  & 0.31  & 6.90 ($\uparrow$) \\
          & BLIP  & 0.26  & 0.28  & 7.69 ($\uparrow$)  \\
          & GIT   & 0.28  & 0.30   & 7.14 ($\uparrow$)  \\
          & Florence & 0.27  & 0.31  & 14.81 ($\uparrow$)  \\
    \bottomrule
    \end{tabular}%
    }
    \caption{BLEU score and text feature dissimilarity of the sequences generated on original and adversarial images.} 
    % The rightmost column is the relative change (\%) over the results of original images.}
  \label{tab:bleu}%
\end{table}%

\begin{figure}[htbp]
\centering
\includegraphics[width=0.45\textwidth]{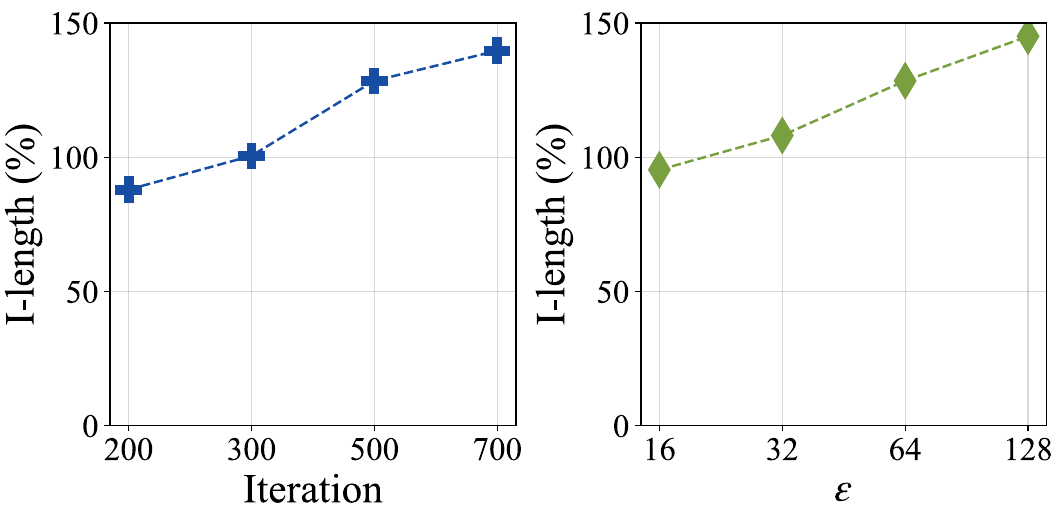}
\caption{Effects of iteration (left) and $L_2$ restriction $\epsilon$ (right). Results are from \textsc{Flamingo} on MS-COCO.}
\label{fig:iterandepsilon}
\end{figure}

\fakeparagraph{Balance Between Efficiency and Unnoticeability.} We study the effect of varying optimization iterations and $L_2$ perturbation restriction. It can be observed in Fig.~\ref{fig:iterandepsilon} that as iteration increases, I-length increases. Similarly, a large $L_2$ restriction results in longer generated sequences. However, more optimization iterations and a larger perturbation lead to more perceptible perturbed images. These parameters balance the \textit{effectiveness} and \textit{unnoticeability} factors mentioned in Sec.~\ref{sec:problemformulation}.

\fakeparagraph{Effect of Defense Method.}
To evaluate whether \tool\ can bypass existing defense mechanisms, we employ Quantization~\cite{xu2017feature} as a defense method. As shown in Tab.~\ref{tab:defense}, neither the efficiency nor the accuracy is significantly impacted by the defense method, demonstrating that Quantization is ineffective against \tool.

% Table generated by Excel2LaTeX from sheet 'defense'
\begin{table}[htbp]
  \centering
    \resizebox{0.475\textwidth}{!}{
    \begin{tabular}{c|ccc|rr}
    \toprule
    \multirow{2}[2]{*}{\textbf{Defense}} & \multicolumn{3}{c|}{\textbf{Efficiency}} & \multicolumn{2}{c}{\textbf{Accuracy}} \\
          & \textbf{I-length} & \textbf{I-latency} & \textbf{I-energy} & \multicolumn{1}{c}{\textbf{BLEU}} & \multicolumn{1}{c}{\textbf{dissim.}} \\
    \midrule
    w/o defense & 128.47       &  105.56     &  115.19     &   0.15    & 0.31 \\
    w/ Quantization & 124.36      &  101.07     &  110.25     & 0.15      & 0.31 \\
    \bottomrule
    \end{tabular}%
    }
    \caption{Efficiency and accuracy metrics for \textsc{Flamingo} on MS-COCO with and without Quantization defense.}
  \label{tab:defense}%
\end{table}%

\begin{figure}[htbp]
\centering
\includegraphics[width=0.475\textwidth]{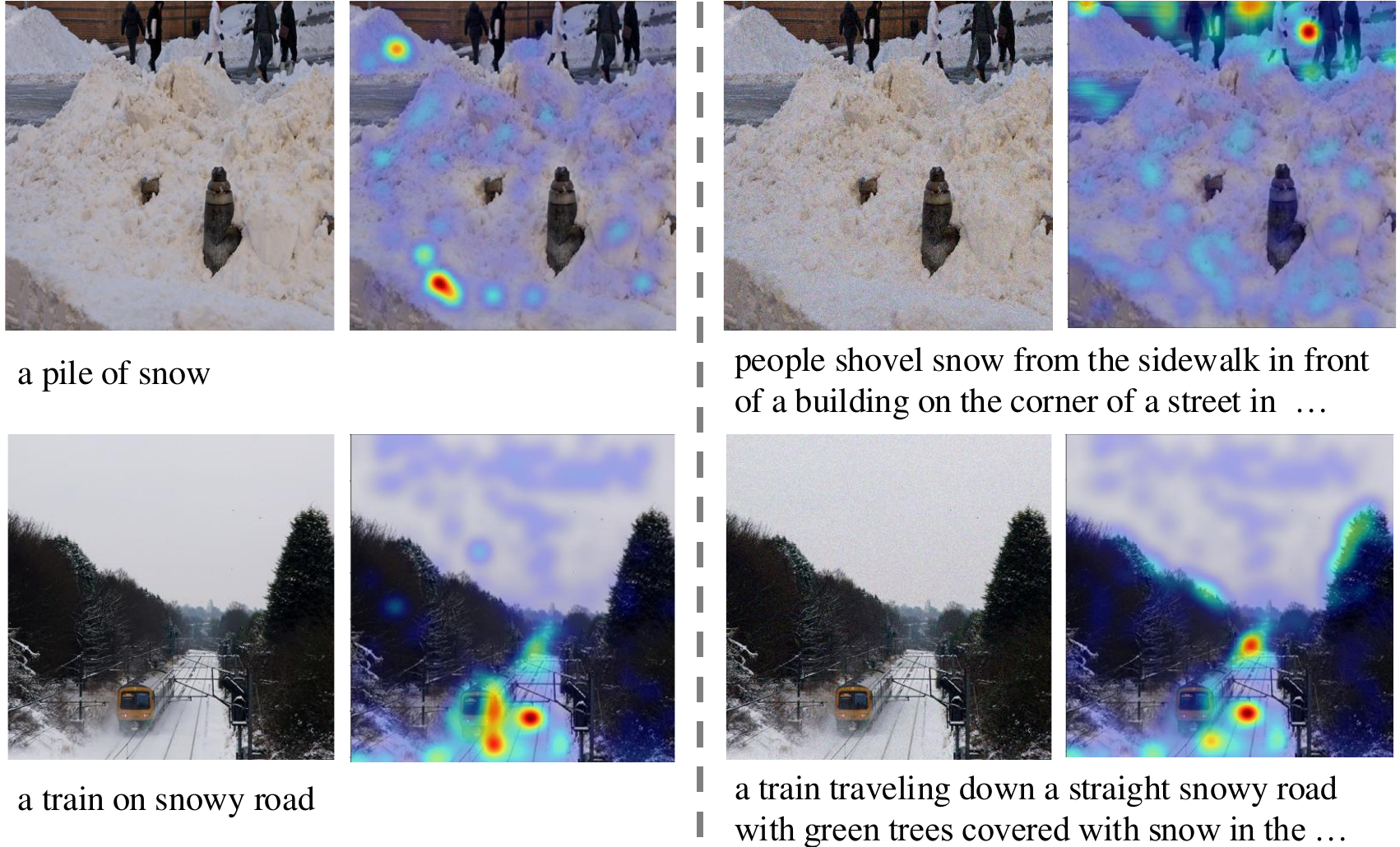}
\caption{Examples of original images (left) and adversarial images (right), and their GradCAM visualizations.}
\label{fig:gradcam}
\end{figure}

\fakeparagraph{Visual Interpretation.}
In Fig.~\ref{fig:gradcam}, we employ GradCAM~\cite{selvaraju2017grad} to visualize image regions that contribute to sequence generation for both original and adversarial images. It demonstrates that for original images, VLM mainly focuses on regions containing objects relevant to the generated sequence, while for adversarial images, the attention maps are dispersed across the entire image. 
This suggests that the longer sequence generated for adversarial images may be attributed to the dispersed attention for visual inputs.

Other discussions, such as robustness to different sampling strategies, are presented in Appendix \ref{sec:appendix_addition}.

% \vspace{-3mm}
\section{Conclusion}

In this paper, 
%we present the first systematic evaluation of the efficiency robustness of VLMs under the black-box setting. 
we introduce \tool, a novel black-box attack framework designed to evaluate the efficiency robustness of VLMs. 
%Specifically, we formulate three objectives to approximate resource consumption and propose a derivative-free gradient estimation algorithm to search for adversarial images. 
Extensive experiments demonstrate that \tool\ significantly increases the computational cost across four VLMs during inference while generating imperceptible adversarial perturbed images. We hope this work raises the community's awareness about the efficiency robustness of VLMs.

\section*{Limitations}
We pioneer an efficiency attack approach under the black-box setting. However, our work has limitations. Firstly, the \tool\ attack requires more optimization iterations compared with white-box approaches, which restricts its effectiveness in scenarios with limited request quotas within a given timeframe.

% Second, our approach assumes the VLM API provides textual outputs and logits. In scenarios where the logits are inaccessible (i.e., only the output sequence is returned), our approach is not directly applicable. We leave this as a direction for future research.

Secondly, our work primarily focuses on developing an efficiency attack approach for VLMs, with limited exploration of defense strategies. We hope future research will propose more robust algorithms to defend against such efficiency attacks, particularly in the black-box setting. This would enhance the trustworthiness and security of VLMs in real-world applications.

\section*{Ethical Considerations}
We acknowledge that our proposed efficiency attack could potentially be exploited for malicious purposes. However, our goal is not to enable such actions but rather to reveal the efficiency vulnerabilities in vision-language models that have been largely overlooked and to raise awareness within the research community. By doing so, we aim to motivate the development of robust defenses against such attacks. We are committed to ethical research and firmly oppose any harmful or unethical use of our findings.

\section*{Acknowledgement}
This work is supported by the Research Grants Council (RGC) of Hong Kong, SAR, China (GRF-16308321), the NSFC/RGC Joint Research Scheme Grant N\_HKUST635/20, and the Beijing Jiaotong University ``Jingying Plan'' No. K24XKRC00130.

% \section*{Acknowledgments}

% This document has been adapted
% by Steven Bethard, Ryan Cotterell and Rui Yan
% from the instructions for earlier ACL and NAACL proceedings, including those for
% ACL 2019 by Douwe Kiela and Ivan Vuli\'{c},
% NAACL 2019 by Stephanie Lukin and Alla Roskovskaya,
% ACL 2018 by Shay Cohen, Kevin Gimpel, and Wei Lu,
% NAACL 2018 by Margaret Mitchell and Stephanie Lukin,
% Bib\TeX{} suggestions for (NA)ACL 2017/2018 from Jason Eisner,
% ACL 2017 by Dan Gildea and Min-Yen Kan,
% NAACL 2017 by Margaret Mitchell,
% ACL 2012 by Maggie Li and Michael White,
% ACL 2010 by Jing-Shin Chang and Philipp Koehn,
% ACL 2008 by Johanna D. Moore, Simone Teufel, James Allan, and Sadaoki Furui,
% ACL 2005 by Hwee Tou Ng and Kemal Oflazer,
% ACL 2002 by Eugene Charniak and Dekang Lin,
% and earlier ACL and EACL formats written by several people, including
% John Chen, Henry S. Thompson and Donald Walker.
% Additional elements were taken from the formatting instructions of the \emph{International Joint Conference on Artificial Intelligence} and the \emph{Conference on Computer Vision and Pattern Recognition}.

% % Bibliography entries for the entire Anthology, followed by custom entries
% %\bibliography{anthology,custom}
% % Custom bibliography entries only
\bibliography{custom}

% \appendix

% \section{Example Appendix}
% \label{sec:appendix}

% This is an appendix.

\clearpage
\appendix

\section{Background}
\subsection{Natural Efficiency Variance}
\begin{figure}[htbp]
\centering
\includegraphics[width=0.475\textwidth]{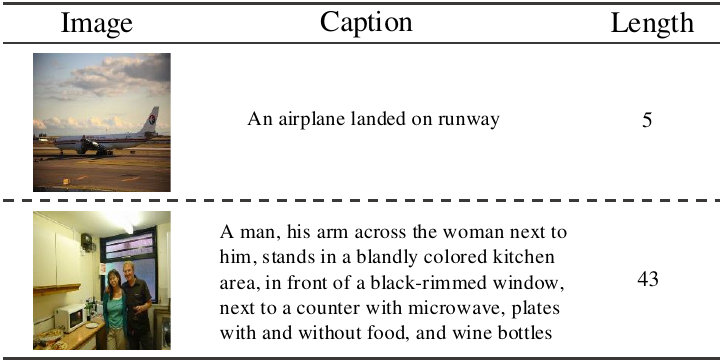}
\caption{Images in MS-COCO with different lengths of captions.}
% Variability for images from MS-COCO.
\label{fig:nev}
\end{figure}

VLMs terminate token generation if the end-of-sequence (EOS) token is sampled or the maximum sequence length is reached. However, presetting an optimal maximum length is challenging due to the inherent variability in the semantic content of different images, as illustrated in Fig.~\ref{fig:nev}. As a result, the common practice is to set a sufficiently large value for the maximum length to avoid generating truncated sequences. 
% Our proposed attack focuses on prolonging the generated sequence, thereby reducing the inference efficiency of VLMs.

\section{Algorithm}
\label{sec:algorithm}

We provide the complete procedure of our \tool\ approach as follows.

\begin{algorithm}[htbp]
\caption{\textbf{VLMInferSlow}}
\textbf{Input}: Benign image $\mathcal{I}$, victim model $\mathcal{F}(\cdot)$, optimization iteration $T$, number of sampled Gaussian noise $q$, search variance $\eta$, update step size $\gamma$, maximum perturbation $\epsilon$   \\
\textbf{Output}: Adversarial perturbation $\delta$ 

\begin{algorithmic}[1]
\State $\delta=0$ \Comment{Initialize perturbation}
\For{$\text{iter}=1$ \textbf{to} $T$}
    \State $\hat{g}=0$ \Comment{Initialize gradient $\hat{g}$}
    \For{$i=1$ \textbf{to} $q$} \Comment{Estimate gradient}
        \State $\mu_i=\mathcal{N}(0, I)$ 
        % \Comment{Sample Gaussian noise}
        \State $\delta_{+}=\delta+\eta\mu_i$; \ \  $\delta_{-}=\delta-\eta\mu_i$ 
        \State $\hat{g}_i=\frac{1}{2\eta}[\mathcal{L}(\delta_+)-\mathcal{L}(\delta_-)]\mu_i$  \Comment{Eq.~\ref{eq:totol_loss},~\ref{eq:gradient}}
        \State $\hat{g}=\hat{g}+\frac{1}{q}\hat{g}_i$ 
    \EndFor
    \State $\delta=\delta+\gamma\times\hat{g}$
    % /\|\hat{g}\|$ 
    \Comment{Update perturbation}
    \State $\delta = \text{Clip}(\delta,\epsilon)$ 
    \Comment{Clip perturbation}
\EndFor
\State\Return $\delta$
\end{algorithmic}
\label{algorithm}
\end{algorithm}

\section{Implementation Details}
\label{sec:appendix_modelanddetails}

\subsection{Evaluated Datasets and Metrics}
\fakeparagraph{Datasets.}
Following~\citet{gaoinducing}, We evaluate our approach on the image captioning task using MS-COCO~\cite{lin2014microsoft} and ImageNet~\cite{deng2009imagenet}. From each dataset, we randomly select 1000 images for evaluation.

\fakeparagraph{Test Hardware.} Following~\citet{gaoinducing} and ~\citet{chen2022nicgslowdown}, we use I-length, I-latency, and I-energy as the evaluated metrics. Latency and energy depend on the hardware. We clarify that all metrics are measured on a single NVIDIA GeForce RTX 3090 GPU.

\fakeparagraph{Definition of Feature Dissimilarity.}
In Sec.~\ref{sec:qualityofgeneratedimages} and Sec.~\ref{sec:morestudies}, we use the image feature dissimilarity and text feature dissimilarity as the distance metrics. Here we clarify the formal definition as follows. Given two features $f_1$ and $f_2$ (image features extracted by CLIP~\cite{radford2021learning} or text features extracted by BERT~\cite{devlin2018bert}), their dissimilarity is:
\begin{equation*}
dissimilarity(f_1,f_2)=1-\frac{f_1f_2}{\|f_1\|\|f_2\|}.
\end{equation*}
In our results, we calculate the dissimilarity between the feature of the original image (or corresponding generated sequence) and the feature of the adversarial counterpart.

\subsection{Target Models}
In our work, we employ four  VLMs as our target victim models: \textsc{Flamingo}~\cite{alayrac2022flamingo}, \textsc{Blip}~\cite{li2022blip}, \textsc{Git}~\cite{wang2022git}, and \textsc{Florence}~\cite{xiao2024florence}. The details of these models are elaborated as follows. All the models are open-sourced and can be downloaded from HuggingFace.

\fakeparagraph{Settings for \textsc{Flamingo}.} Suggested by~\citet{alayrac2022flamingo}, we employ the CLIP ViT-large-patch14 and OPT-125M LM. The image resolution is 224$\times$224, and a placeholder $\emptyset$ is taken as the prompt text $\mathcal{T}$.

\fakeparagraph{Settings for \textsc{Blip}.} We use the \textsc{Blip} with the basic multimodal mixture
of encoder-decoder in the 224M version. Following~\citet{li2022blip}, for our image captioning task, the image resolution is 384$\times$384, and a placeholder $\emptyset$ is taken as the prompt text $\mathcal{T}$.

\fakeparagraph{Settings for \textsc{Git}.} We utilize the base-sized version of the \textsc{Git} model which has been fine-tuned on MS-COCO. The image resolution is 224$\times$224, and the prompt text $\mathcal{T}$ is a placeholder $\emptyset$, following~\citet{wang2022git}.

\fakeparagraph{Settings for \textsc{Florence}.} We choose the base-sized \textsc{Florence-2 SD3}~\cite{xiao2024florence} as the victim model.
The image resolution is 224$\times$224, and we use the default text prompt:
\textit{<DESCRIPTION> Describe this image} for the image captioning task.

\subsection{Optimization and Setting Details}
% \fakeparagraph{Sequence Generation Process.} We use greedy search for the sequence generation process for the experimental results shown in our main paper. We also present the results of using different sampling strategies in the Sec.~\ref{sec:appendix_addition}.

\fakeparagraph{Optimization Details.} 
We provide more details of our approach. The maximum perturbation is set as $\epsilon=64$. 
As for the parameters of the sampled gaussian noise stated in Sec.~\ref{sec:gradientestimation}, we set $q=5$ and $\eta=0.1$. For the sequence generation process, we adopt greedy search as the sampling strategy. 
% Following~\citet{gaoinducing}, we randomly select 1000 images in each dataset for evaluation.

% we adopt the nucleus sampling strategy with $p=0.9$ and temperature $temp=1$. 

\fakeparagraph{Black-box Evaluation Setting Details.} We assume that the model architecture and parameters are inaccessible. In our baselines, we compare our \tool\  with two white-box methods (NICGSlowdown and Verbose Images) and two natural image corruptions (Gaussian and JPEG). We measure the performance of two white-box methods as follows. For one model (e.g., \textsc{Flamingo}), we test the performance of the perturbed images optimized on the three other models (e.g., \textsc{Blip}, \textsc{Git}, and \textsc{Florence}). 
% For black-box methods (\tool\, Gaussian, and JPEG), we directly report the result of the perturbed images optimized on the tested model itself.

\fakeparagraph{Results of White-box Baselines.} We also compare \tool\ in the black-box setting with the results of two baselines NICGSlowdown and Verbose Images in the white-box settings.
% For the white-box methods, we report the results of the perturbed images optimized on the test model itself, assuming that the whole model, including the architecture and parameters, is accessible. 
We run for 50 iterations for two white-box methods.

\section{Derivation of Gradient Estimation}
\label{sec:appendix_derivation}
As stated in Sec.~\ref{sec:gradientestimation}, we maximize the expected value of the objective $\mathcal{L}(\cdot)$ under a search distribution $\pi(z|\delta)$: 
\begin{equation*}
    J(\delta) = \mathbb{E}_{\pi(z|\delta)}[\mathcal{L}(z)]=\int \mathcal{L}(z)\pi(z|\delta)\ dz.
\end{equation*}
Then, the gradient $\nabla_{\delta}J(\delta)$ can be computed as:
\begin{equation*}
\begin{aligned}
\nabla_{\delta}J(\delta) &= \nabla_{\delta}\int \mathcal{L}(z)\pi(z|\delta)\ dz\\
         &=\int\mathcal{L}(z)\nabla_{\delta}\ \pi(z|\delta)\ dz \\
         &=\int\mathcal{L}(z)\frac{\nabla_{\delta}\ \pi(z|\delta)}{\pi(z|\delta)}\pi(z|\delta)\ dz \\
         &= \int[\mathcal{L}(z)\nabla_{\delta}\log \pi(z|\delta)] \pi(z|\delta)\ dz \\
         &= \mathbb{E}_{\pi(z|\delta)}[\mathcal{L}(z)\nabla_{\delta}\log\pi(z|\delta)]
\end{aligned}
\end{equation*}
We obtain $z_i$ by $z_i=\delta+\eta\mu_i$, $i\in\{1,2,\cdots,2q\}$, where $\mu_i$ is the gaussian noise sampled from $\mathcal{N}(0,I)$. 
Then, $\pi(z|\delta)$ is a normal distribution $\mathcal{N}(\delta,\eta^2I)$. In this way, we have:
\begin{equation*}
    \nabla_{\delta}\log\pi(z|
\delta) = \frac{z-\delta}{\eta^2} = \frac{\mu}{\eta}.
\end{equation*}
Thus, the gradient $\nabla_{\delta}J(\delta)$ can be estimated as:
\begin{equation*}
\hat{\nabla}_{\delta}J(\delta)=
% \approx
\frac{1}{2\eta q}\sum_{i=1}^{2q}\mu_i\mathcal{L}(\delta+\eta\mu_i).
\end{equation*}

\section{Additional Results}
\label{sec:appendix_addition}

% \begin{table*}[htbp]
%   \centering
%   \small
%     \begin{tabular}{ccc|cccccc}
%     \toprule
%     \multicolumn{3}{c|}{\textbf{losses}} & \multicolumn{3}{c}{\textbf{MS-COCO}} & \multicolumn{3}{c}{\textbf{ImageNet-1k}} \\
%     \multicolumn{1}{c}{$\mathcal{L}_{len}$} & \multicolumn{1}{c}{$\mathcal{L}_{eos}$} & \multicolumn{1}{c|}{$\mathcal{L}_{var}$} & \multicolumn{1}{c}{\textbf{I-length}} & \multicolumn{1}{c}{\textbf{I-latency}} & \multicolumn{1}{c}{\textbf{I-energy}} & \multicolumn{1}{c}{\textbf{I-length}} & \multicolumn{1}{c}{\textbf{I-latency}} & \multicolumn{1}{c}{\textbf{I-energy}} \\
%     \midrule
%     \checkmark     &       &       & 28.59 & 18.21 & 32.49 & 32.96 & 30.23 & 38.59 \\
%           & \checkmark     &       & 42.15 & 27.01 & 45.65 & 53.99 & 44.01 & 47.73 \\
%           &       & \checkmark     & 23.22 & 13.09  & 30.15 & 32.96 & 30.12 & 33.80 \\
%     \checkmark     & \checkmark     &       & 67.65 & 45.51 & 60.41 & 70.32 & 61.43 & 50.71 \\
%     \checkmark     &       & \checkmark     & 54.77 & 36.72 & 57.59 & 49.58 & 40.01 & 37.65 \\
%           & \checkmark     & \checkmark     & 42.68 & 27.41 & 50.09 & 53.32 & 45.21 & 45.90 \\
%     \checkmark     & \checkmark     & \checkmark    & \textbf{71.95} & \textbf{54.98} & \textbf{65.41} & \textbf{74.38} & \textbf{66.89} & \textbf{55.58} \\
%     \bottomrule
%     \end{tabular}%
% \caption{Ablation results of three designed losses. Results are on \textsc{BLIP}.}
% \label{tab:ablation_loss_total}
% \end{table*}%

\subsection{Comparison with White-box Methods}
Due to page limit, in Sec.~\ref{sec:mainresults}, we only provide the results for two white-box baselines of MS-COCO in Tab.~\ref{tab:whitebox}. Here we provide the results of ImageNet-1k as in Tab.~\ref{tab:whitebox_in1k}. We can find that even if optimized in a black-box setting, our \tool\ achieves comparable performance with white-box methods which assume total access to the target model. This aligns with our observation in the main paper.

\begin{table}[htbp]
  \centering
\resizebox{0.475\textwidth}{!}{
    \begin{tabular}{ccccc}
    \toprule
    \textbf{Models} & \textbf{Methods} & \textbf{I-length} & \textbf{I-latency} & \textbf{I-energy} \\
    \midrule
    \multirow{3}[1]{*}{Flamingo} & NICGSlowdown &      51.35 & 36.84 & 43.82  \\
          & Verbose &  \textbf{137.92} & \textbf{105.26} & \textbf{101.15}  \\
          & \tool &    103.44 & 78.42 & 70.32 \\
          \midrule
    \multirow{3}[0]{*}{BLIP} & NICGSlowdown &   71.61 & 70.14 & 81.48  \\
          & Verbose & \textbf{134.35} & \textbf{121.24} & \textbf{105.72} \\
          & \tool &   74.38 & 66.89 & 55.58 \\      
          \midrule
    \multirow{3}[0]{*}{GIT} & NICGSlowdown &74.05 & 61.54 &78.26 \\
          & Verbose & \textbf{119.21}  &\textbf{138.46}  &\textbf{107.86} \\
          & \tool &     {93.86} & {115.38} & {84.82}  \\          \midrule
    \multirow{3}[1]{*}{Florence} & NICGSlowdown &      37.07 & 41.21 & 51.59\\
          & Verbose & 23.86 & 23.21 & 50.92 \\
          & \tool &  \textbf{47.39} & \textbf{42.86} & \textbf{65.83}  \\
    \bottomrule
    \end{tabular}%
}
\caption{Comparision with two white-box baselines on ImageNet-1k. Best results are in \textbf{bold}.}
\label{tab:whitebox_in1k}
\end{table}%

\begin{table*}[t]
  \centering
\small
   \resizebox{0.85\textwidth}{!}{
    \begin{tabular}{cccccccc}  
    \toprule
    \multirow{2}[2]{*}{\textbf{Models}} & \multirow{2}[2]{*}{\textbf{Methods}} & \multicolumn{3}{c}{\textbf{MS-COCO}} & \multicolumn{3}{c}{\textbf{ImageNet-1k}} \\
          &       & \textbf{I-length} & \textbf{I-latency} & \textbf{I-energy} & \textbf{I-length} & \textbf{I-latency} & \textbf{I-energy} \\
    \midrule
    \multirow{5}[1]{*}{LLaVA} & Gaussian & 8.76  & 1.85  & 0.88  & 9.12  & 3.91  & 14.77 \\
          & JPEG  & -12.87 & -1.03 & -4.73 & 12.38 & 10.82 & 7.93 \\
          & NICGSlowdown-B & -8.31 & -0.96 & -3.42 & 3.32  & 2.48  & -0.73 \\
          & Verbose-B & 1.41  & 0.11  & 2.32  & 2.14  & 1.67  & 0.48 \\
          & \tool & \textbf{78.56} & \textbf{81.29} & \textbf{63.48} & \textbf{82.47} & \textbf{70.21} & \textbf{67.64} \\
    \midrule
    \multirow{5}[1]{*}{Qwen} & Gaussian & 5.62  & 1.18  & 10.89 & -10.53 & -4.23 & -1.33 \\
          & JPEG  & 3.21  & 4.56  & 1.78  & -2.43 & -4.69 & -1.91 \\
          & NICGSlowdown-B & -0.23 & -0.78 & -2.67 & -0.55 & -0.57 & -2.72 \\
          & Verbose-B & 2.59  & 1.74  & 2.52  & 1.91  & 0.88  & 3.97 \\
          & \tool & \textbf{61.32} & \textbf{57.95} & \textbf{49.90} & \textbf{67.94} & \textbf{60.82} & \textbf{52.10} \\
    \midrule
    \multirow{5}[1]{*}{MiniGPT} & Gaussian & -15.13 & -2.34 & -12.58 & -12.17 & -18.39 & -17.42 \\
          & JPEG  & 11.71 & 8.63  & 3.44  & 9.34  & 2.55  & 5.32 \\
          & NICGSlowdown-B & 4.56  & 3.78  & 0.24  & 0.31  & 6.66  & 1.78 \\
          & Verbose-B & 3.95  & 1.33  & 5.91  & 1.83  & 0.43  & 3.45 \\
          & \tool & \textbf{47.41} & \textbf{64.98} & \textbf{49.57} & \textbf{56.09} & \textbf{42.77} & \textbf{50.13} \\
    \bottomrule
    \end{tabular}%
    }
    \caption{Results on more VLMs. Best results are in \textbf{bold}.}
  \label{tab:more_vlms}%
\end{table*}%

% \subsection{Additional Ablation Studies}
\subsection{Hyperparameter Sensitivity}
\begin{figure}[htbp]
\centering
\includegraphics[width=0.475\textwidth]{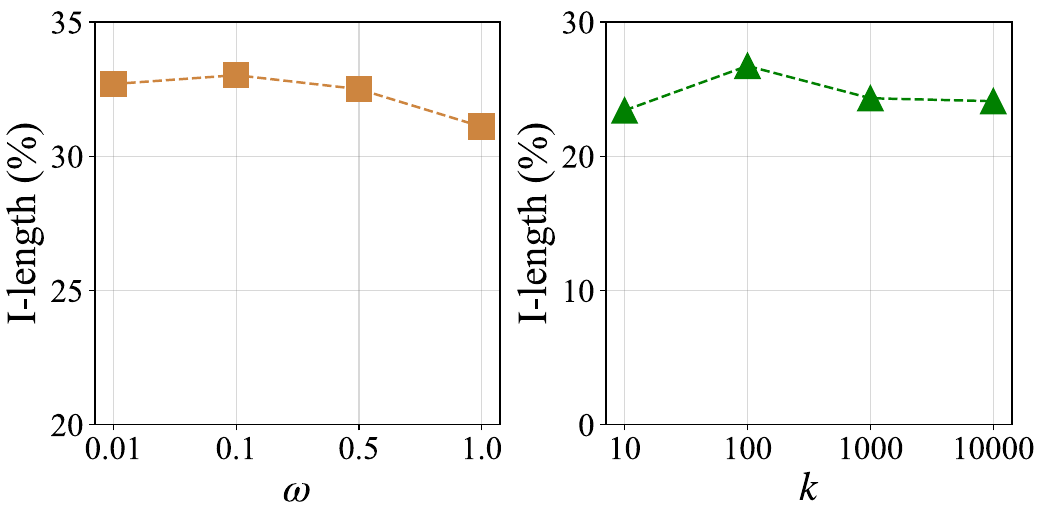}
\caption{Effects of $\omega$ in $\mathcal{L}_{eos}$ (left) and $k$ in $\mathcal{L}_{var}$ (right). Results are from \textsc{Flamingo} on MS-COCO.}
\label{fig:wandk}
\end{figure}

\fakeparagraph{Effect of $\omega$ in $\mathcal{L}_{eos}$.}
We adopt the dynamic weight decay strategy in $\mathcal{L}_{eos}$,
% \begin{equation*}
% \mathcal{L}_{eos} =  \sum_{i=1}^{N}\omega^{N-i}\text{Pr}^{\text{EOS}}(y_i),
% \end{equation*}
in which $\omega$ controls the weight decay speed. We study the parameter sensitivity of $\omega$. As shown in Fig.~\ref{fig:wandk}, it can be observed that the optimal performance is achieved when $w$ is set as 0.1.  When $w=1.0$ (i.e., the average of the EOS token probabilities across all positions is used), performance slightly declines. This validates the efficacy of our dynamic weight decay strategy.

\fakeparagraph{Effect of $k$ in $\mathcal{L}_{var}$.}
In $\mathcal{L}_{var}$, we select the top-$k$ probabilities in $\text{Pr}(\cdot)$ and normalize them to form a new probability distribution $\Tilde{\text{Pr}}(\cdot)$. We set $k=100$ in practice since the large size of the vocabulary can make the objective difficult to optimize. To investigate the impact of $k$, we vary k under $[10,100,1000,10000]$ in Fig.~\ref{fig:wandk}. It can be observed that when $k=100$, the optimal result is yield, and further increasing $k$ provides no additional performance improvement.

\subsection{Effect of Different Sampling Strategies} 
In our main results, we use the greedy search sampling strategy
% nucleus sampling strategy
for sequence generation. We further investigate the impact of different sampling strategies, including nucleus sampling, top-k sampling, and beam search. As shown in Tab.~\ref{tab:samplingstrat}, \tool\ consistently generates longer sequences across all scenarios, demonstrating its effectiveness in reducing the efficiency of VLMs under different sequence generation policies.

\begin{table}[htbp]
  \centering
\resizebox{0.475\textwidth}{!}{
    \begin{tabular}{cccc}
    \toprule
    \textbf{Strategies} & \textbf{I-length} & \textbf{I-latency} & \textbf{I-energy} \\
    \midrule
    Greedy search &  128.47     & 105.56      & 115.19 \\
    Beam search &   123.58    &   93.83    & 109.57 \\
    Top-k sampling  &     113.30  & 85.22      &  93.98\\
   Nucleus sampling &  130.42     &   111.95    & 120.74 \\
    \bottomrule
    \end{tabular}%
    }
    \caption{Results of different sampling strategies against \textsc{Flamingo} on MS-COCO.}
  \label{tab:samplingstrat}%
\end{table}%

\subsection{Results of More VLMs}
We evaluate the effectiveness of \tool\ on more VLMs, including LLaVA1.5~\cite{liu2023visual}, Qwen2.5-VL~\cite{bai2025qwen2}, and MiniGPT4~\cite{zhu2023minigpt}. The experimental setup is the same as in the main paper.  Results are presented in Tab.~\ref{tab:more_vlms}. It shows that for all VLMs, \tool\ significantly outperforms baselines for decreasing efficiency of VLMs in the black-box evaluation setting, which aligns with the conclusion in our main paper.

\section{Visualization}
\label{sec:appendix_visualization}
\begin{figure*}[t]
\centering
\includegraphics[width=0.9\textwidth]{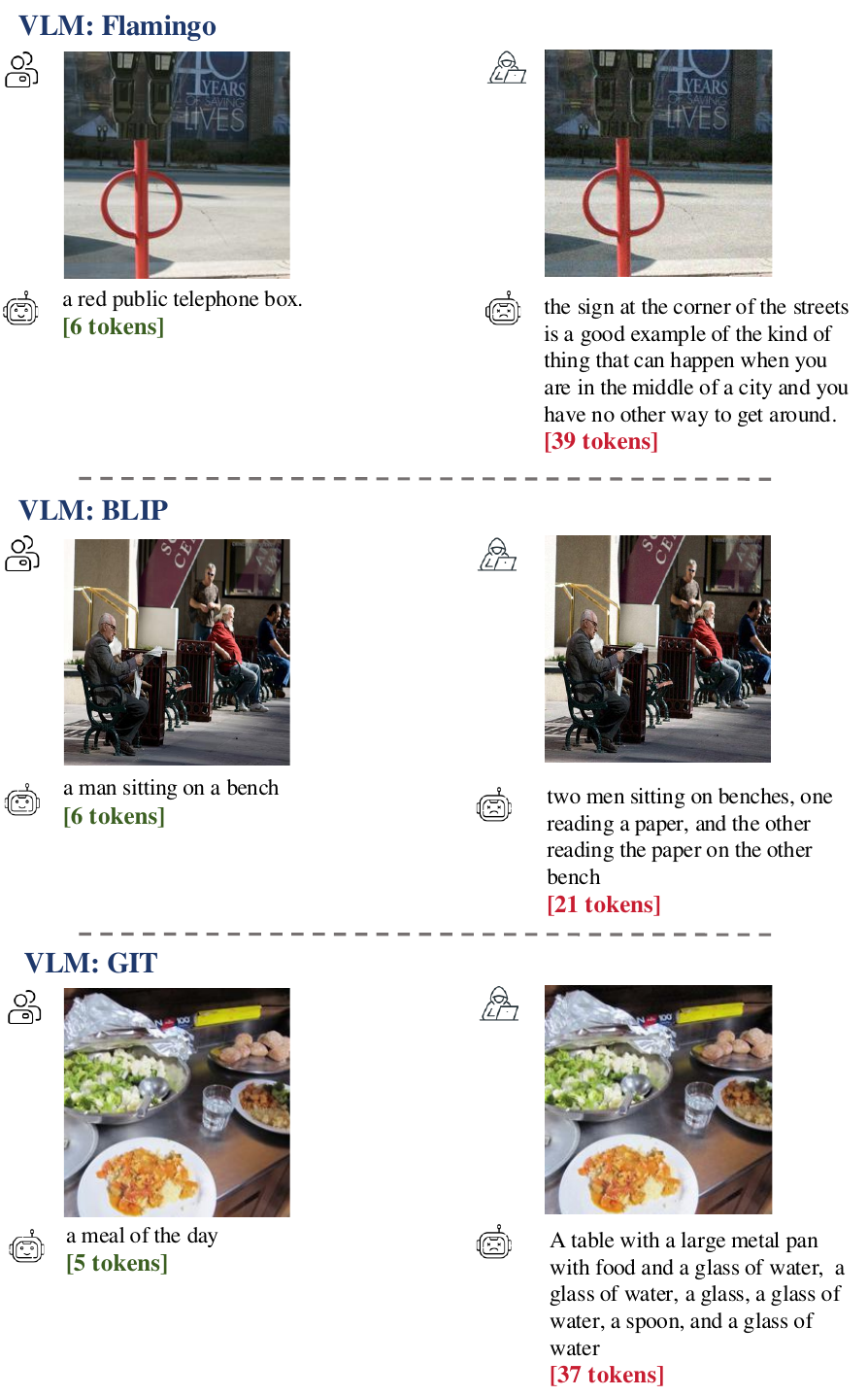}
\caption{Visualization of original images (left) and adversarial images (right) generated by \tool\ against Flamingo, BLIP, and GIT, along with their corresponding output sequences.}
\label{fig:example}
\end{figure*}

\begin{figure*}[t]
\centering
\includegraphics[width=0.9\textwidth]{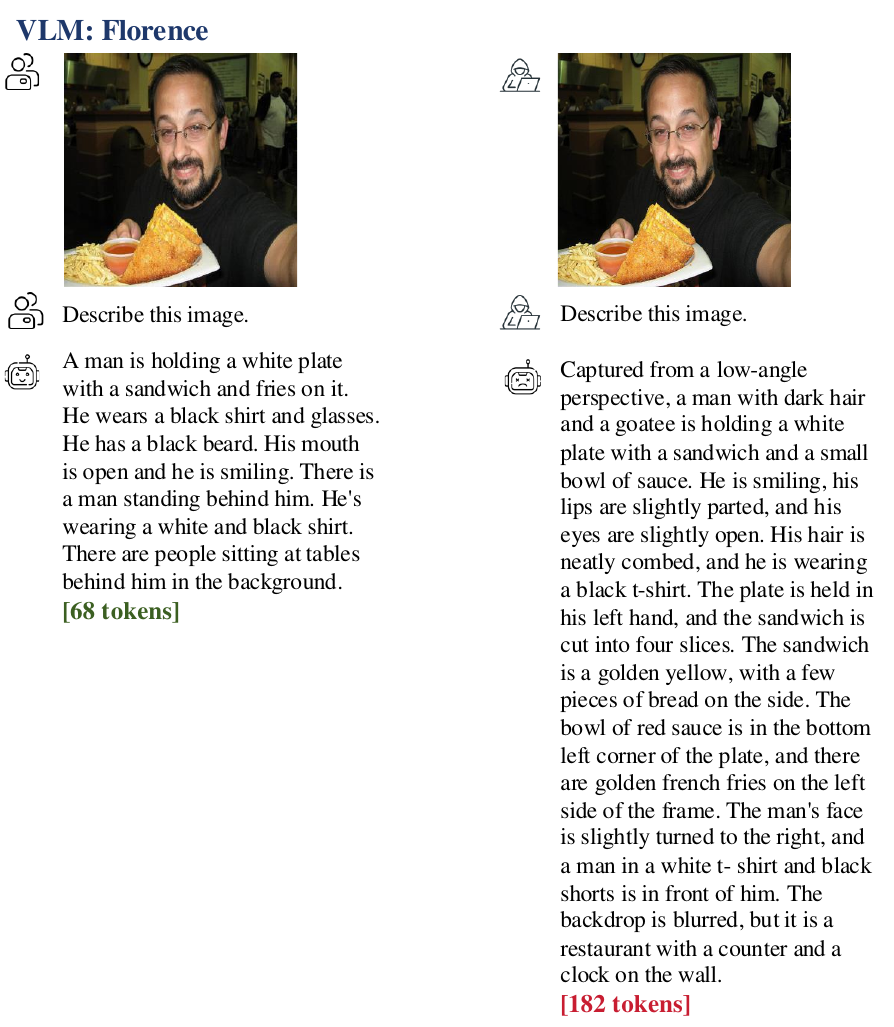}
\caption{Visualization of an original image (left) and an adversarial image (right) generated by our proposed \tool\ against Florence, along with its corresponding output sequence.}
\label{fig:example1}
\end{figure*}

In this section, we present examples of the original and adversarial images generated by \tool, along with their corresponding generated sequences against four different VLMs. The figures and output sequences are presented in Fig.~\ref{fig:example} and Fig.~\ref{fig:example1}. It can be observed that the perturbations added to adversarial images are imperceptible to human observers, and all four VLMs generate longer sequences on adversarial images than original images.

% Several observations can be made: (1) the perturbations added by \tool\ are imperceptible to human observers; (2) all four VLMs generate longer sequences for adversarial images than original images, resulting in increased response time and higher energy consumption, as discussed in our main paper; (3) \tool\ is applicable to diverse VLMs and settings. For instance, \textsc{Flamingo}, which typically generates concise sequences for original images, produces longer and more detailed sequences for adversarial images. Furthermore, while \textsc{Florence} already generates long sequences for original images, it produces even more lengthy sequences when processing adversarial images.

\end{document}